\documentclass{article}

    \usepackage[nonatbib,final]{neurips_2023}

\usepackage[utf8]{inputenc} 
\usepackage[T1]{fontenc}    
\usepackage{hyperref}       
\usepackage{url}            
\usepackage{booktabs}       
\usepackage{amsfonts}       
\usepackage{nicefrac}       
\usepackage{microtype}      
\usepackage{xcolor}         

\usepackage{doi}
\usepackage{tikz}
\usepackage{svg}

\usepackage{caption}
\usepackage{graphicx}
\usepackage{bbm}
\usepackage{wrapfig}
\usepackage{amsthm}
\usepackage{enumitem}
\usepackage{relsize}
\usepackage{pdflscape}

\usepackage{multicol, multirow, xltabular}

\usepackage{amsmath}
\usepackage{booktabs}

\usepackage{amsmath,amsfonts,bm}

\def\1{\bm{1}}

\DeclareMathAlphabet{\mathsfit}{\encodingdefault}{\sfdefault}{m}{sl}
\SetMathAlphabet{\mathsfit}{bold}{\encodingdefault}{\sfdefault}{bx}{n}

\usepackage{todonotes}

\usepackage{subcaption}

\renewcommand{\paragraph}[1]{\textbf{#1}\hspace{1em}}

\makeatletter
\newcommand{\printfnsymbol}[1]{%
  \footnotemark[#1]\hspace{1.5mm}%
}
\makeatother

\title{Self-Distilled Representation Learning for Time Series} 

\author{%
  Felix Pieper\thanks{Equal contribution}\hspace{1.5mm}\thanks{Corresponding author (\url{felix.pieper@merantix.com})}\hspace{1.5mm}\printfnsymbol{4}, Konstantin Ditschuneit\printfnsymbol{1}\thanks{Work done while at Merantix Momentum}\hspace{1.5mm}, Martin Genzel\printfnsymbol{1}\printfnsymbol{4}, \\ \bfseries Alexandra Lindt\printfnsymbol{1}\printfnsymbol{3}, and Johannes Otterbach\printfnsymbol{5}\printfnsymbol{3} \\
  \printfnsymbol{4}Merantix Momentum, \printfnsymbol{5}nyonic\\
}

\begin{document}

\maketitle

\begin{abstract}
    Self-supervised learning for time-series data holds potential similar to that recently unleashed in Natural Language Processing and Computer Vision.
    While most existing works in this area focus on contrastive learning, we propose a conceptually simple yet powerful non-contrastive approach, based on the data2vec self-distillation framework.
    The core of our method is a student-teacher scheme that predicts the latent representation of an input time series from masked views of the same time series. 
    This strategy avoids strong modality-specific assumptions and biases typically introduced by the design of contrastive sample pairs. 
    We demonstrate the competitiveness of our approach for classification and forecasting as downstream tasks, comparing with state-of-the-art self-supervised learning methods on the UCR and UEA archives as well as the ETT and Electricity datasets.
\end{abstract}

\section{Introduction}
\label{sec:intro}

Time series are a ubiquitous data resource in numerous application domains, ranging from finance and healthcare to environmental monitoring and manufacturing. Understanding and harnessing their inherent patterns is the driving force behind standard tasks like predictive analytics, forecasting, and anomaly detection. Despite a notable body of work on \emph{deep learning techniques} for time-series analysis~\cite{survey-ts-classification,survey-ts-forecasting,survey-ts-transformers}, more traditional methods like XGBoost~\cite{xgboost} and handcrafted features continue to play a pivotal role, often setting the gold standard in supervised learning and forecasting~\cite{forecasting-trees,nixtla-study,transformers-ineffective}.
In particular, learning universal representations remains a fundamental challenge for temporal data.

Given recent breakthroughs in Natural Language Processing (NLP) and Computer Vision (CV), the paradigm of \emph{self-supervised learning (SSL)} has the potential to become a game changer in the area of time series as well. 
While huge amounts of unlabeled temporal data exist in many business sectors, it is fair to say that research in this direction and practical feasibility are still not mature.

A popular line of work on SSL for time series focuses on \emph{contrastive methods}, aiming at robust data representations by training neural networks to differentiate between positive (similar) and negative (dissimilar) pairs of samples.
Few prominent examples are TS2Vec~\cite{ts2vec}, T-Loss~\cite{tloss}, TS-TCC~\cite{ts-tcc}, and TNC~\cite{tnc}; see \cite{survey-pretrain,survey-ssl} for recent surveys of the field.
Although the effectiveness of contrastive methods has been demonstrated on several benchmark datasets, e.g., see~\cite{survey-pretrain}, the design of positive and negative samples for time series is not straightforward. Indeed, common augmentation strategies from CV and NLP are not easily transferable, as modality-specific characteristics like temporal and multi-scale dependencies need to be considered.
As a consequence, the performance of existing methods often strongly depends on the specific use case and task.

\emph{Non-contrastive methods} are a promising remedy for addressing this lack of flexibility.
As the name suggests, this class of algorithms focuses on pretext tasks that encourage a model to learn meaningful data representations without the explicit construction of positive and negative sample pairs.
While there exists a variety of non-contrastive SSL approaches in CV and NLP, e.g., see~\cite{dino,byol,data2vec,simsiam,vicreg,barlow}, they found much less attention in the time-series domain. In fact, existing work mostly focuses on ``classical'' unsupervised learning techniques like autoencoders~\cite{ti-mae,timemae,timenet}, see~\cite{survey-pretrain,survey-ssl} for a broader overview.

The present work makes further progress on non-contrastive learning for time-series data and presents, to the best of our knowledge, the first method based on self-distillation.
Our main contributions can be summarized as follows:
\begin{enumerate}
\item
    We propose a conceptually simple non-contrastive learning strategy for time-series data by adopting the recent \emph{data2vec} framework~\cite{data2vec}. 
    The underlying idea is to leverage a student-teacher scheme to predict the \emph{latent} representation of given input data based on a masked view of the same input.
    Unlike contrastive methods, no modality-specific designs are required in this process.
    On a larger scope, we underpin the main promise of data2vec (originally considered for vision, language, and speech) to provide a seamlessly extendable, modality-agnostic framework.
\item 
    We demonstrate the effectiveness of our method for classification and forecasting downstream tasks. In comparison with several existing SSL approaches, we report state-of-the-art performance on the UCR~\cite{ucr}~\& UEA~\cite{uea} benchmark archives for classification as well as on the ETT~\cite{ett_informer}~\& Electricity~\cite{electricity} datasets for forecasting.
\end{enumerate}

\section{Methodology}
\label{sec:method}

\begin{figure}
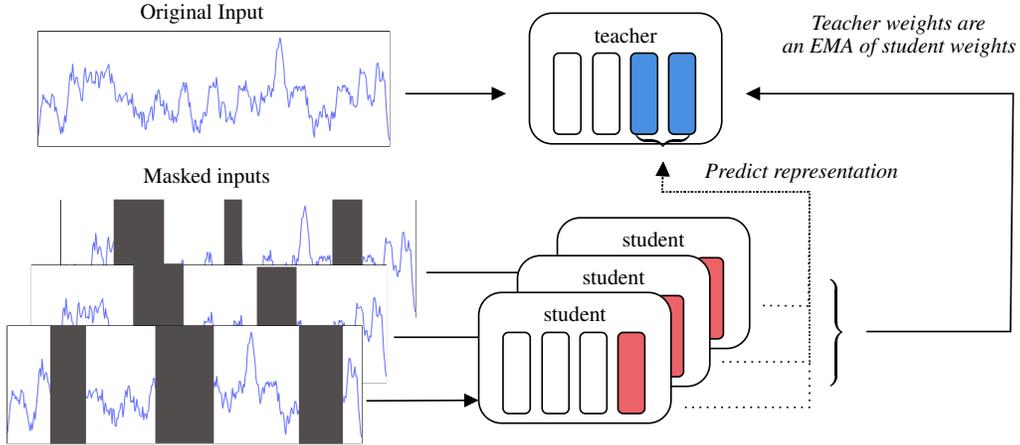

    \centering
    \include{figures/train_objective_timeseries}
    \vspace{-1\baselineskip}
    \caption{\emph{Illustration of our data2vec-based training pipeline.} In teacher mode, the encoder computes a target representation of the full input time series by averaging the hidden activations of the last $K$ encoder layers (shaded in blue). 
    In student mode, this representation is then predicted by encoding (multiple) versions of the same input with randomly masked timestamps (shaded in red). As common in self-distillation schemes, the teacher's weights follow the student's weights according to an EMA.}
    \label{fig:train_objective}
\end{figure}

\paragraph{Self-distillation training objective.}
Our training strategy closely follows the SSL approach of \emph{data2vec}~\cite{data2vec}, which proposes a simple, yet effective self-distillation scheme.
The \emph{teacher model} provides a target representation of given input data, which the \emph{student model} is supposed to predict from masked versions of the same input; see Figure~\ref{fig:train_objective} for an illustration.
More specifically, the target representation is computed by averaging the hidden activations over the last $K$ layers of the teacher model, which was found to stabilize the training dynamics~\cite{data2vec}.
Similarly to related self-distillation frameworks like BYOL~\cite{byol} or DINO~\cite{dino}, the teacher's weights follow the student model according to an \emph{Exponential Moving Average (EMA)} mechanism during training.
We argue that data2vec is well-suited for our purposes because of its simplicity and generalizability. 
Our simple timestamp masking strategy particularly bypasses the limitations and unintentional biases that typically occur when handcrafting positive and negative samples in contrastive methods.

For a more detailed introduction to data2vec and an in-depth analysis of its design choices, we refer to the original paper~\cite{data2vec} as well as its successor data2vec~2.0~\cite{data2vec2}.
Moreover, we point out some differences between our approach and the original framework in Appendix~\ref{app:sec:exp}.

\paragraph{Encoder architecture.}
A notable difference from the original data2vec scheme is our choice of encoder backbone: instead of a transformer-based model, we employ a \emph{Convolutional Neural Network (CNN)}.
This design choice aligns our approach closer with existing (contrastive) SSL methods for time-series data and allows for a more direct comparison.
Our specific architecture is inspired by the \emph{TS2Vec} encoder~\cite{ts2vec}, which proposes a cascade of residual convolutional blocks.
Here, the $l$-th block applies two consecutive 1D convolutions with dilation parameter $2^l$ to enlarge the receptive field over the temporal axis.
At the same time, a suitable padding scheme ensures consistent feature dimensionality from layer to layer, which is key to the hierarchical contrastive loss function developed in~\cite{ts2vec}.
Our learning protocol also exploits this consistency albeit in a different way, namely by computing the averaged target representation vector over the last $K$ layers.
In contrast to the original TS2Vec architecture, we have incorporated batch normalization after each convolutional layer as well as a tunable scaling factor for the weight initialization, both of which enhanced the stability of our self-distillation training pipeline.

Finally, let us emphasize that the representations produced by our CNN encoder are still sequential, i.e., one feature vector is computed per timestamp.
While this is analogous to transformer-based encoders, our feature vectors get ``contextualized'' by exponentially increasing the dilation parameter instead of using self-attention layers.

\section{Experimental Results}
\label{sec:experiments}

We assess the effectiveness of our method with respect to its downstream task performance in time-series classification and forecasting.
Our basic experimental setup follows a simple two-step procedure for each considered dataset: (1) learning the encoder in a self-supervised fashion without any labels, and (2) training a task-specific layer on top of the learned representations, while the encoder's weights are frozen.
This protocol is closely aligned with the one of TS2Vec~\cite{ts2vec}, which will serve as our primary reference point for comparison with state-of-the-art SSL methods;\footnote{The reported scores for all comparison methods---not only TS2Vec---are taken from~\cite{ts2vec}; see Appendix~\ref{app:sec:methods} for more details.} see also~\cite{survey-pretrain} for an independent benchmarking study.

It is well-known that self-distillation is prone to representation collapse, which is why we performed a preliminary hyperparameter optimization (HPO) on a small subset of the UCR archive to select important training parameters like the learning rate or EMA parameters. In the actual experiments, all hyperparameters (including the CNN encoder architecture, which is not tuned) remain fixed and consistent.
For more details on the experimental setup and implementation, see Appendix~\ref{app:sec:exp}.

\paragraph{Time-series classification.} In this standard downstream task, each time-series instance is associated with a single label to be predicted. To obtain instance-level representations, we first perform a max-aggregation over all timestamps. The resulting (fixed-size) feature vector is then used as input for an SVM head with RBF kernel, which is trained on the labeled dataset.
Following~\cite{ts2vec}, we benchmark our approach on the UCR archive~\cite{ucr} and UEA archive~\cite{uea}, which consist of 128 (univariate) and 30 (multivariate) datasets, respectively.

\begin{table}[t]
\centering
\caption{Summarized results for time-series classification using accuracy as metric. We report the average scores over all datasets of each archive ($128$ datasets for UCR and $30$ datasets for UEA, respectively). See Table~\ref{tab:ucr} and~\ref{tab:uea} in Appendix~\ref{app:sec:results} for the full results.} \vspace{.5\baselineskip}
\label{tab:classification_mean}
\resizebox{\textwidth}{!}{%
\begin{tabular}{lcccccccc}
\toprule
 & Ours & TS2Vec \cite{ts2vec} & Ti-MAE \cite{ti-mae} & T-Loss \cite{tloss} & TNC \cite{tnc} & TS-TCC \cite{ts-tcc} & TST \cite{tst} \\
\midrule
UCR & \textbf{0.832} & 0.829 & 0.823 & 0.806 & 0.761 & 0.757 & 0.638 \\
UEA & \textbf{0.738} & 0.704 & - & 0.658 & 0.670 & 0.668 & 0.617 \\
\bottomrule
\end{tabular}
}
\end{table}

Our experimental results are summarized in Table~\ref{tab:classification_mean}. For the UCR archive, we have also included the scores reported for Ti-MAE~\cite{ti-mae}, which is a recent non-contrastive approach based on a masked autoencoder.
We conclude that our data2vec scheme is highly competitive with existing SSL methods, slightly outperforming TS2Vec on UCR and even more clearly on UEA. Here, the UEA archive can be considered more challenging due to its multivariate nature.

\paragraph{Time-series forecasting.} Given a time series up to a certain timestamp, forecasting aims to predict future observations.
Our downstream protocol first extracts the last encoded feature vector (corresponding to the last observed timestamp), which is then used as input to train a ridge regression head that predicts the next $H$ observations.
Adopting the experimental setup of~\cite{ts2vec} again, we consider three versions of the ETT datasets~\cite{ett_informer} as well as the Electricity dataset~\cite{electricity}, both in the uni- and multivariate setting.

Our results are summarized in Table~\ref{tab:forecast_uni_mean}.
While our data2vec approach is consistently competitive in the univariate case, we highlight a notable improvement in MSE on Electricity and ETTh$_2$.
Similarly to classification, the performance gain becomes even more striking in the multivariate case, for which we report superior results across almost all datasets. 

\begin{table}[t]
\centering
\caption{Summarized results for time-series forecasting using mean squared error (MSE) and mean absolute error (MAE) as metrics. For each dataset, we report the average scores over all values of~$H$ (= number of future observations to be predicted). See Table~\ref{tab:forecast_uni} and~\ref{tab:forecast_multi} in Appendix~\ref{app:sec:results} for the full results, including more comparison methods.} \vspace{.5\baselineskip}
\label{tab:forecast_uni_mean}
\resizebox{\textwidth}{!}{%
\begin{tabular}{clcccccccccc}
\toprule
& & \multicolumn{2}{c}{Ours} & \multicolumn{2}{c}{TS2Vec \cite{ts2vec}} & \multicolumn{2}{c}{Informer \cite{ett_informer}} & \multicolumn{2}{c}{LogTrans \cite{logtrans}} & \multicolumn{2}{c}{TCN \cite{tcn}} \\
\cmidrule(lr){3-4}  \cmidrule(lr){5-6}  \cmidrule(lr){7-8}  \cmidrule(lr){9-10}  \cmidrule(lr){11-12}
& Dataset & MSE & MAE  & MSE & MAE  & MSE & MAE  & MSE & MAE  & MSE & MAE \\
\midrule
\multirow{5}{*}{\rotatebox[origin=c]{90}{Univariate}} & ETTh$_1$ & 0.1303 & 0.2744 & \textbf{0.1104} & \textbf{0.2524} & 0.186 & 0.3468 & 0.196 & 0.3646 & 0.2628 & 0.4314 \\
& ETTh$_2$ & \textbf{0.1445} & \textbf{0.2944} & 0.1698 & 0.321 & 0.204 & 0.3582 & 0.2174 & 0.391 & 0.2186 & 0.3616 \\
& ETTm$_1$ & 0.0741 & 0.1952 & \textbf{0.069} & \textbf{0.1864} & 0.2412 & 0.382 & 0.2702 & 0.4164 & 0.1998 & 0.3488 \\
& Electricity & \textbf{0.3263} & \textbf{0.4243} & 0.4864 & 0.4246 & 0.6072 & 0.4712 & 0.7952 & 0.5652 & 0.6726 & 0.5098 \\
\cmidrule(lr){2-12}
& Avg. & \textbf{0.1688} & 0.2971 & 0.209 & \textbf{0.296} & 0.31 & 0.39 & 0.37 & 0.434 & 0.338 & 0.413 \\
\midrule
\multirow{5}{*}{\rotatebox[origin=c]{90}{Multivariate}} 
& ETTh1 & \textbf{0.667} & \textbf{0.616} & 0.788& 0.646& 0.907& 0.739& 1.043& 0.89& 1.021& 0.816\\
& ETTh2 & \textbf{0.716} & \textbf{0.65} & 1.567& 0.937& 2.371& 1.199& 2.898& 1.356& 2.574& 1.265\\
& ETTm1 & \textbf{0.506} & \textbf{0.522} & 0.628& 0.553& 0.749& 0.64& 0.965& 0.914& 0.818& 0.849\\
& Electricity & \textbf{0.297} & \textbf{0.392} & 0.33& 0.405& 0.589& 0.548& 0.35& 0.41 & 0.355& 0.42\\ 
\cmidrule(lr){2-12}
& Avg.& \textbf{0.546} & \textbf{0.545} & 0.828& 0.636& 1.154& 0.781& 1.314& 0.892 & 1.192& 0.837\\
\bottomrule
\end{tabular}
}
\end{table}

\section{Discussion}
\label{sec:conclusion}

This work provides initial evidence for the effectiveness of SSL via self-distillation in the time-series domain.
Our experimental study particularly demonstrates that state-of-the-art performance in classification and forecasting is achievable without strong modality-specific assumptions, which are typically made by contrastive methods.

\paragraph{Scope and limitations.} Despite competitive empirical results, the scope of our work is linked to the limitations of the considered benchmark archives.
Although these datasets are diverse and widely used in the related literature, we believe that they are not perfectly suited for an assessment of large-scale (deep-)learning methods, especially SSL and pre-training techniques.
For example, the UCR archive contains rather small datasets, some of which have quite degenerated train/test splits, resulting in noisy and insignificant evaluations regardless of the used learning algorithm.
A specific limitation of our self-distillation framework is its sensitivity to the training parameters. In fact, to produce robust representations and prevent model collapses, additional hyperparameter tuning is required in advance.

\paragraph{Outlook and challenges.} Obvious avenues of future research are the exploration of other non-contrastive methods as well as different types of encoder backbones.
In the bigger picture, we argue that large-scale experiments are indispensable to unleash and certify the power of (SSL) deep-learning methods for time-series analysis.
To catch up with the more mature fields of CV and NLP, perhaps the most important challenge is the creation of large, inhomogeneous cohorts of (publicly available) time-series data; see TimeGPT~\cite{timegpt} for a very recent effort in that direction.
Beyond that, we believe that more fundamental modality-specific research is required for future breakthroughs. 
For instance, temporal data still lacks a unified tokenization strategy, unlike NLP and CV where well-established tokenizers are crucial to the current success of Large Language Models and Vision Transformers~\cite{vit}.

\clearpage

\section*{Acknowledgements}
This work has received funding from the German Federal Office for Information Security as part of the EMiL project under grant no.~01MO23014B.
We would like to thank Lisa Coiffard for her constructive feedback in the preparation of this paper.

{\small
\bibliography{main}
\bibliographystyle{abbrv}
}

\appendix
\section{Additional Implementation Details}
\label{app:sec:exp}

This part complements Section~\ref{sec:experiments} and describes more details on the implementation of our experiments as well as some specific design choices.

\paragraph{Datasets.} 
All considered datasets are accompanied by predefined splits, which we adopt to ensure direct comparability with the other methods.

The UCR archive~\cite{ucr} has established itself as a standard benchmark in time-series classification, providing a collection of 128 univariate datasets from various fields such as finance, healthcare, and climatology.
The UEA archive~\cite{uea} is similarly diverse with the important difference that it contains multivariate datasets, and can therefore be seen as more challenging.

The ETT datasets~\cite{ett_informer} provide hourly energy consumption metrics and are widely used as a forecasting benchmark. We consider the versions ETTh$_1$, ETTh$_2$, and ETTm$_1$ in the uni- and multivariate case.
The Electricity dataset~\cite{electricity} records high-frequency household electricity consumption data. Its size and complexity make it a suitable testbed for time-series representation learning techniques as well.

\paragraph{Downstream evaluation.}
For time-series classification, we follow the experimental setup of~\cite{ts2vec}, which is based on the standard SVM implementation of \texttt{scikit-learn}, specified to an RBF kernel and a one-vs-rest strategy for multiclass classification; each evaluation step involves a simple grid search cross-validation to optimize for the SVM regularization parameter \texttt{C}.
We perform downstream evaluations at regular intervals during pre-training to assess the quality of our learned representations.
The validation accuracy of the best evaluation then yields the final performance score. 

Our approach to time-series forecasting is analogous. Here, we use the standard ridge regression module of \texttt{scikit-learn}, tuning the regularization strength \texttt{alpha} through a grid search cross-validation in each downstream evaluation.

\paragraph{Hyperparameters and pre-training.}
Across all experiments, we use the Adam optimizer~\cite{adam} and set the training batch size to $8$.
Following data2vec, we use a Smooth L1-loss to measure the distance between the teacher's representation targets and the student's predictions.

For the pre-training phase, we ensure that each dataset undergoes an equivalent number of time steps. 
This means that the total number of training steps is proportional to the length of the time series. We also include a warmup phase, using the OneCycle learning rate scheduler to prevent overfitting on local minima and to allow sufficient time for the batch normalization to adjust. 
To address the higher dimensionality of the UEA datasets, we crop each input time series to a random window of size $1024$.
These windows are selected independently for each sample and every training iteration. Similarly, for forecasting, we use a cropping window size of $200$, which is consistent with TS2Vec.

To stabilize our self-distillation approach, some important training parameters are selected through a preliminary HPO. As auxiliary validation metric, we apply our representation learning method to $8$~UCR datasets and measure the average accuracy achieved by training a simple logistic regression classification head.
The tuned hyperparameters are as follows: learning rate \& scheduler warm-up parameter, weight decay, EMA parameters, block masking probability, encoder dropout rate, scaling factor for the random weight initialization of the encoder, and the $\beta$-parameter of the Smooth L1-loss.
The selected parameters are used consistently over all experiments described in Section~\ref{sec:experiments}.

All experiments were conducted on a Kubernetes Cluster hosted on Google Cloud Plattform, using NVIDIA Tesla T4 GPU accelerators.
\texttt{PyTorch}~\cite{pytorch} and \texttt{Lightning} are used as underlying deep learning framework.

\paragraph{data2vec and CNN encoder.}
Compared to the original data2vec framework~\cite{data2vec}, we also incorporate some extensions from data2vec~2.0~\cite{data2vec2}.
The first adaptation is a block masking algorithm, which is a simplification of the inverse masking technique proposed by data2vec~2.0. 
Our approach iterates through each (student) batch of time series data until the cumulative proportion of masked blocks marginally exceeds a predefined masking probability. 
In every iteration step, we inject a new masked block into each time series, where the size and location of this block are randomly selected and bounded by the missing blocks. 
This ensures that each time series of a batch has masked blocks that vary in size and position, thereby enhancing the robustness of the representation learning process.
Note that the masking probability was tuned through our preliminary HPO.

A second notable adoption from data2vec~2.0 is the use of multiple student representations to amortize the costs of the teacher model computation.
In our experiments, the number of students is consistently set to $3$. 

We update the teacher weights according to an EMA:
\[
    w_{\text{teacher}} \leftarrow (1-\delta) \cdot w_{\text{student}} + \delta \cdot w_{\text{teacher}}.
\]
The update parameter $\delta$ starts at $0.9996$ and linearly increases to $0.99996$ over the training process. These choices were proposed by our preliminary HPO, but it is noteworthy that the numerical values differ only in the least significant digits from the ones reported in data2vec~\cite{data2vec}.

For the CNN encoder described in Section~\ref{sec:method}, we use $7$ residual convolutional layers, which equals the number of layers $K$ over which the data2vec teacher model computes its representation, i.e., all hidden encoder blocks are used for the averaging. 
The feature dimension in the representation space is set to $320$ (per timestamp), which equals the choice of TS2Vec. 

\section{Comparison Methods}
\label{app:sec:methods}

Below, we briefly describe all comparison methods considered in Section~\ref{sec:experiments}.
Our selection is adopted from \cite{ts2vec}, which is also the origin of the scores reported in this work (except for our method and Ti-MAE~\cite{ti-mae}). We refer to \cite{ts2vec} for reproduction details of each method as well as a more extensive discussion of conceptual differences between them.

Comparison methods for classification:
\begin{itemize}
\item 
    TS2Vec~\cite{ts2vec} proposes a contrastive method that learns contextual representations based on a hierarchical loss that considers contrast on multiple resolution scales. Here, negative samples are obtained both instance-wise and on the temporal axis. Positive samples are generated through contextual consistency of augmented views of the input time series. 
\item 
    Ti-MAE~\cite{ti-mae} introduces a non-contrastive representation learning approach, which randomly masks parts of an embedded time series and learns to reconstruct it through an autoencoder scheme. Both the encoder and decoder are based on transformer blocks.
\item 
    TNC~\cite{tnc} proposes a contrastive learning approach that exploits the local smoothness of time-series signals to define neighborhoods over the temporal axis. Their contrastive loss intends to distinguish the encoded representations of neighborhood signals from non-neighborhood signals.
\item
    TS-TCC~\cite{ts-tcc} leverages both temporal and contextual contrasting, encouraging similarity among different contexts of the same time-series sample while minimizing similarity among contexts of different samples. Weak and strong augmentations are used to generate different yet correlated views.
\item 
    T-Loss~\cite{tloss} proposes a representation learning approach based on a triplet loss and time-based negative sampling. Here, random sub-time series are used to design positive pairs, while different time-series instances are used as negative pairs.
\item 
    TST~\cite{tst} uses a transformer-based model for pre-training on multivariate time series. Their training objective is inspired by BERT-style models \cite{bert}, predicting a time-series signal from a randomly masked version thereof. 
\end{itemize}

Comparison methods for forecasting:
\begin{itemize}
\item 
    Informer~\cite{ett_informer} is a transformer-based model specifically designed for long-sequence time-series forecasting. It proposes an efficient probabilistic attention mechanism achieving $O(L \log L)$ complexity in time and memory, thereby avoiding the well-known quadratic bottleneck of standard attention modules.
\item 
    LogTrans~\cite{logtrans} proposes the LogSparse Transformer architecture, which is based on a convolutional self-attention block that enhances the incorporation of local context. In this way, they achieve super-linear memory complexity, similarly to the Informer.
\item
    The authors of TCN~\cite{tcn} conducted a systematic experimental study of generic convolutional and recurrent architectures for sequence modeling.
    They find that a simple Temporal Convolutional Network (TCN) outperforms common recurrent architectures such as LSTMs on various tasks and datasets.
\item 
    LSTnet~\cite{lstnet} leverages a combination of Convolutional and Recurrent Neural Networks that takes into account both short-term local dependencies and long-term trends in sequential data.
\item 
    N-BEATS~\cite{nbeats} proposes a deep stack of multi-layer fully connected blocks with forward and backward residual connections. A particular feature of their network design is that its outputs are human-interpretable.
\end{itemize}

\section{Full Experimental Results}
\label{app:sec:results}

The full results on all datasets of the UCR and UEA archive are reported in Table~\ref{tab:ucr} and~\ref{tab:uea}, respectively.
Table~\ref{tab:forecast_uni} and~\ref{tab:forecast_multi} show the full results of our forecasting experiments in the uni- and multivariate case, respectively.

\clearpage

{\newgeometry{left=2.5cm}
\begin{xltabular}{\textwidth}{lccccccc}
\caption{Full results for time-series classification on the \textbf{UCR} archive using accuracy as metric. The reported scores for Ti-MAE are taken from~\cite{ti-mae} and the scores for all other comparison methods are taken from~\cite{ts2vec}. Note that for each method, pre-training and the downstream task are performed for each dataset individually.} \label{tab:ucr} \\
\toprule
UCR Dataset & Ours & TS2Vec & Ti-MAE & T-Loss & TNC & TS-TCC & TST  \\
\midrule
\endfirsthead
\multicolumn{8}{c}{\tablename\ \thetable{}: Full results for time-series classification on the \textbf{UCR} archive (continued from previous page).} \\ \\
\toprule
UCR Dataset & Ours & TS2Vec & Ti-MAE & T-Loss & TNC & TS-TCC & TST \\
\midrule
\endhead
\\ \multicolumn{8}{c}{(continued on next page)}
\endfoot
\bottomrule
\endlastfoot
Avg. acc. & \textbf{0.832} & 0.829 & 0.823 & 0.806 & 0.761 & 0.757 & 0.638 \\
Avg. rank & \textbf{2.109} & 2.586 & 2.656 & 3.720 & 4.584 & 4.512 & 6.234 \\
\midrule
ACSF1 & 0.830 & \textbf{0.900} & 0.820 & \textbf{0.900} & 0.730 & 0.730 & 0.760 \\
Adiac & \textbf{0.788} & 0.762 & \textbf{0.788} & 0.675 & 0.726 & 0.767 & 0.550 \\
AllGestureWiimoteX & 0.776 & \textbf{0.777} & 0.633 & 0.763 & 0.703 & 0.697 & 0.259 \\
AllGestureWiimoteY & 0.777 & \textbf{0.793} & 0.682 & 0.726 & 0.699 & 0.741 & 0.423 \\
AllGestureWiimoteZ & \textbf{0.749} & 0.746 & 0.671 & 0.723 & 0.646 & 0.689 & 0.447 \\
ArrowHead & 0.857 & 0.857 & \textbf{0.874} & 0.766 & 0.703 & 0.737 & 0.771 \\
BME & \textbf{1.000} & 0.993 & \textbf{1.000} & 0.993 & 0.973 & 0.933 & 0.760 \\
Beef & 0.667 & 0.767 & \textbf{0.900} & 0.667 & 0.733 & 0.600 & 0.500 \\
BeetleFly & 0.950 & 0.900 & 0.900 & 0.800 & 0.850 & 0.800 & \textbf{1.000} \\
BirdChicken & 0.800 & 0.800 & \textbf{1.000} & 0.850 & 0.750 & 0.650 & 0.650 \\
CBF & 0.997 & \textbf{1.000} & \textbf{1.000} & 0.983 & 0.983 & 0.998 & 0.898 \\
Car & 0.850 & 0.833 & \textbf{0.867} & 0.833 & 0.683 & 0.583 & 0.550 \\
Chinatown & 0.965 & 0.965 & \textbf{0.985} & 0.951 & 0.977 & 0.983 & 0.936 \\
ChlorineConcentration & 0.754 & \textbf{0.832} & 0.725 & 0.749 & 0.760 & 0.753 & 0.562 \\
CinCECGTorso & 0.654 & 0.827 & \textbf{0.971} & 0.713 & 0.669 & 0.671 & 0.508 \\
Coffee & \textbf{1.000} & \textbf{1.000} & \textbf{1.000} & \textbf{1.000} & \textbf{1.000} & \textbf{1.000} & 0.821 \\
Computers & 0.772 & 0.660 & \textbf{0.780} & 0.664 & 0.684 & 0.704 & 0.696 \\
CricketX & 0.767 & \textbf{0.782} & 0.674 & 0.713 & 0.623 & 0.731 & 0.385 \\
CricketY & \textbf{0.751} & 0.749 & 0.659 & 0.728 & 0.597 & 0.718 & 0.467 \\
CricketZ & 0.754 & \textbf{0.792} & 0.718 & 0.708 & 0.682 & 0.713 & 0.403 \\
Crop & \textbf{0.763} & 0.756 & 0.751 & 0.722 & 0.738 & 0.742 & 0.710 \\
DiatomSizeReduction & 0.987 & 0.984 & 0.984 & 0.984 & \textbf{0.993} & 0.977 & 0.961 \\
DistalPhalanxOutlineAgeGroup & 0.755 & 0.727 & \textbf{0.763} & 0.727 & 0.741 & 0.755 & 0.741 \\
DistalPhalanxOutlineCorrect & \textbf{0.804} & 0.761 & 0.793 & 0.775 & 0.754 & 0.754 & 0.728 \\
DistalPhalanxTW & \textbf{0.748} & 0.698 & 0.727 & 0.676 & 0.669 & 0.676 & 0.568 \\
DodgerLoopDay & \textbf{0.613} & 0.562 & \textbf{0.613} & -- & -- & -- & 0.200 \\
DodgerLoopGame & \textbf{0.913} & 0.841 & 0.739 & -- & -- & -- & 0.696 \\
DodgerLoopWeekend & \textbf{0.978} & 0.964 & \textbf{0.978} & -- & -- & -- & 0.732 \\
ECG200 & \textbf{0.940} & 0.920 & 0.910 & \textbf{0.940} & 0.830 & 0.880 & 0.830 \\
ECG5000 & 0.940 & 0.935 & \textbf{0.942} & 0.933 & 0.937 & 0.941 & 0.928 \\
ECGFiveDays & \textbf{1.000} & \textbf{1.000} & 0.988 & \textbf{1.000} & 0.999 & 0.878 & 0.763 \\
EOGHorizontalSignal & \textbf{0.608} & 0.539 & 0.558 & 0.605 & 0.442 & 0.401 & 0.373 \\
EOGVerticalSignal & 0.475 & 0.503 & \textbf{0.547} & 0.434 & 0.392 & 0.376 & 0.298 \\
Earthquakes & \textbf{0.748} & \textbf{0.748} & \textbf{0.748} & \textbf{0.748} & \textbf{0.748} & \textbf{0.748} & \textbf{0.748} \\
ElectricDevices & 0.704 & \textbf{0.721} & 0.685 & 0.707 & 0.700 & 0.686 & 0.676 \\
EthanolLevel & 0.670 & 0.468 & \textbf{0.744} & 0.382 & 0.424 & 0.486 & 0.260 \\
FaceAll & 0.873 & 0.771 & \textbf{0.880} & 0.786 & 0.766 & 0.813 & 0.504 \\
FaceFour & 0.795 & \textbf{0.932} & 0.875 & 0.920 & 0.659 & 0.773 & 0.511 \\
FacesUCR & 0.908 & \textbf{0.924} & 0.866 & 0.884 & 0.789 & 0.863 & 0.543 \\
FiftyWords & 0.769 & 0.771 & \textbf{0.787} & 0.732 & 0.653 & 0.653 & 0.525 \\
Fish & \textbf{0.937} & 0.926 & 0.897 & 0.891 & 0.817 & 0.817 & 0.720 \\
FordA & 0.930 & \textbf{0.936} & 0.818 & 0.928 & 0.902 & 0.930 & 0.568 \\
FordB & 0.790 & 0.794 & 0.652 & 0.793 & 0.733 & \textbf{0.815} & 0.507 \\
FreezerRegularTrain & \textbf{0.998} & 0.986 & 0.987 & 0.956 & 0.991 & 0.989 & 0.922 \\
FreezerSmallTrain & 0.980 & 0.870 & 0.959 & 0.933 & \textbf{0.982} & 0.979 & 0.920 \\
Fungi & 0.989 & 0.957 & 0.968 & \textbf{1.000} & 0.527 & 0.753 & 0.366 \\
GestureMidAirD1 & 0.654 & 0.608 & \textbf{0.662} & 0.608 & 0.431 & 0.369 & 0.208 \\
GestureMidAirD2 & \textbf{0.631} & 0.469 & 0.546 & 0.546 & 0.362 & 0.254 & 0.138 \\
GestureMidAirD3 & 0.331 & 0.292 & \textbf{0.400} & 0.285 & 0.292 & 0.177 & 0.154 \\
GesturePebbleZ1 & 0.738 & \textbf{0.930} & 0.901 & 0.919 & 0.378 & 0.395 & 0.500 \\
GesturePebbleZ2 & 0.677 & 0.873 & \textbf{0.918} & 0.899 & 0.316 & 0.430 & 0.380 \\
GunPoint & \textbf{1.000} & 0.980 & 0.993 & 0.980 & 0.967 & 0.993 & 0.827 \\
GunPointAgeSpan & \textbf{0.994} & 0.987 & \textbf{0.994} & \textbf{0.994} & 0.984 & \textbf{0.994} & 0.991 \\
GunPointMaleVersusFemale & \textbf{1.000} & \textbf{1.000} & 0.997 & 0.997 & 0.994 & 0.997 & \textbf{1.000} \\
GunPointOldVersusYoung & \textbf{1.000} & \textbf{1.000} & \textbf{1.000} & \textbf{1.000} & \textbf{1.000} & \textbf{1.000} & \textbf{1.000} \\
Ham & 0.733 & 0.714 & \textbf{0.800} & 0.724 & 0.752 & 0.743 & 0.524 \\
HandOutlines & 0.916 & 0.922 & 0.919 & 0.922 & \textbf{0.930} & 0.724 & 0.735 \\
Haptics & 0.464 & \textbf{0.526} & 0.484 & 0.490 & 0.474 & 0.396 & 0.357 \\
Herring & 0.641 & 0.641 & \textbf{0.656} & 0.594 & 0.594 & 0.594 & 0.594 \\
HouseTwenty & \textbf{0.941} & 0.916 & \textbf{0.941} & 0.933 & 0.782 & 0.790 & 0.815 \\
InlineSkate & \textbf{0.471} & 0.415 & 0.380 & 0.371 & 0.378 & 0.347 & 0.287 \\
InsectEPGRegularTrain & \textbf{1.000} & \textbf{1.000} & \textbf{1.000} & \textbf{1.000} & \textbf{1.000} & \textbf{1.000} & \textbf{1.000} \\
InsectEPGSmallTrain & \textbf{1.000} & \textbf{1.000} & \textbf{1.000} & \textbf{1.000} & \textbf{1.000} & \textbf{1.000} & \textbf{1.000} \\
InsectWingbeatSound & 0.590 & 0.630 & \textbf{0.639} & 0.597 & 0.549 & 0.415 & 0.266 \\
ItalyPowerDemand & 0.963 & 0.925 & \textbf{0.967} & 0.954 & 0.928 & 0.955 & 0.845 \\
LargeKitchenAppliances & \textbf{0.861} & 0.845 & 0.787 & 0.789 & 0.776 & 0.848 & 0.595 \\
Lightning2 & \textbf{0.902} & 0.869 & 0.836 & 0.869 & 0.869 & 0.836 & 0.705 \\
Lightning7 & 0.808 & \textbf{0.863} & 0.808 & 0.795 & 0.767 & 0.685 & 0.411 \\
Mallat & 0.950 & 0.914 & \textbf{0.956} & 0.951 & 0.871 & 0.922 & 0.713 \\
Meat & \textbf{0.967} & 0.950 & \textbf{0.967} & 0.950 & 0.917 & 0.883 & 0.900 \\
MedicalImages & \textbf{0.803} & 0.789 & 0.771 & 0.750 & 0.754 & 0.747 & 0.632 \\
MelbournePedestrian & 0.958 & \textbf{0.959} & 0.949 & 0.944 & 0.942 & 0.949 & 0.741 \\
MiddlePhalanxOutlineAgeGroup & 0.649 & 0.636 & \textbf{0.675} & 0.656 & 0.643 & 0.630 & 0.617 \\
MiddlePhalanxOutlineCorrect & \textbf{0.852} & 0.838 & 0.811 & 0.825 & 0.818 & 0.818 & 0.753 \\
MiddlePhalanxTW & \textbf{0.623} & 0.584 & \textbf{0.623} & 0.591 & 0.571 & 0.610 & 0.506 \\
MixedShapes & \textbf{0.922} & 0.917 & \textbf{0.922} & 0.905 & 0.911 & 0.855 & 0.879 \\
MixedShapesSmallTrain & \textbf{0.877} & 0.861 & 0.875 & 0.860 & 0.813 & 0.735 & 0.828 \\
MoteStrain & 0.880 & 0.861 & \textbf{0.913} & 0.851 & 0.825 & 0.843 & 0.768 \\
NonInvasiveFetalECGThorax1 & 0.924 & \textbf{0.930} & 0.918 & 0.878 & 0.898 & 0.898 & 0.471 \\
NonInvasiveFetalECGThorax2 & 0.930 & \textbf{0.938} & \textbf{0.938} & 0.919 & 0.912 & 0.913 & 0.832 \\
OSULeaf & 0.806 & \textbf{0.851} & 0.736 & 0.760 & 0.723 & 0.723 & 0.545 \\
OliveOil & 0.867 & 0.900 & \textbf{0.933} & 0.867 & 0.833 & 0.800 & 0.800 \\
PLAID & 0.449 & \textbf{0.561} & 0.458 & 0.555 & 0.495 & 0.445 & 0.419 \\
PhalangesOutlinesCorrect & \textbf{0.834} & 0.809 & 0.772 & 0.784 & 0.787 & 0.804 & 0.773 \\
Phoneme & 0.266 & \textbf{0.312} & 0.229 & 0.276 & 0.180 & 0.242 & 0.139 \\
PickupGestureWiimoteZ & 0.700 & 0.820 & \textbf{0.840} & 0.740 & 0.620 & 0.600 & 0.240 \\
PigAirwayPressure & \textbf{0.793} & 0.630 & 0.240 & 0.510 & 0.413 & 0.380 & 0.120 \\
PigArtPressure & 0.904 & \textbf{0.966} & 0.760 & 0.928 & 0.808 & 0.524 & 0.774 \\
PigCVP & \textbf{0.889} & 0.812 & 0.750 & 0.788 & 0.649 & 0.615 & 0.596 \\
Plane & \textbf{1.000} & \textbf{1.000} & \textbf{1.000} & 0.990 & \textbf{1.000} & \textbf{1.000} & 0.933 \\
PowerCons & 0.961 & 0.961 & \textbf{1.000} & 0.900 & 0.933 & 0.961 & 0.911 \\
ProximalPhalanxOutlineAgeGroup & \textbf{0.883} & 0.834 & 0.863 & 0.844 & 0.854 & 0.839 & 0.854 \\
ProximalPhalanxOutlineCorrect & 0.883 & \textbf{0.887} & 0.876 & 0.859 & 0.866 & 0.873 & 0.770 \\
ProximalPhalanxTW & 0.824 & 0.824 & \textbf{0.829} & 0.771 & 0.810 & 0.800 & 0.780 \\
RefrigerationDevices & 0.571 & 0.589 & \textbf{0.611} & 0.515 & 0.565 & 0.563 & 0.483 \\
Rock & \textbf{0.840} & 0.700 & 0.660 & 0.580 & 0.580 & 0.600 & 0.680 \\
ScreenType & 0.480 & 0.411 & \textbf{0.579} & 0.416 & 0.509 & 0.419 & 0.419 \\
SemgHandGenderCh2 & 0.900 & \textbf{0.963} & 0.838 & 0.890 & 0.882 & 0.837 & 0.725 \\
SemgHandMovementCh2 & 0.713 & \textbf{0.860} & 0.700 & 0.789 & 0.593 & 0.613 & 0.420 \\
SemgHandSubjectCh2 & 0.813 & \textbf{0.951} & 0.813 & 0.853 & 0.771 & 0.753 & 0.484 \\
ShakeGestureWiimoteZ & 0.900 & \textbf{0.940} & 0.900 & 0.920 & 0.820 & 0.860 & 0.760 \\
ShapeletSim & \textbf{1.000} & \textbf{1.000} & 0.911 & 0.672 & 0.589 & 0.683 & 0.489 \\
ShapesAll & 0.855 & \textbf{0.902} & 0.840 & 0.848 & 0.788 & 0.773 & 0.733 \\
SmallKitchenAppliances & 0.699 & 0.731 & \textbf{0.741} & 0.677 & 0.725 & 0.691 & 0.592 \\
SmoothSubspace & \textbf{1.000} & 0.980 & 0.993 & 0.960 & 0.913 & 0.953 & 0.827 \\
SonyAIBORobotSurface1 & \textbf{0.918} & 0.903 & 0.912 & 0.902 & 0.804 & 0.899 & 0.724 \\
SonyAIBORobotSurface2 & 0.858 & 0.871 & \textbf{0.934} & 0.889 & 0.834 & 0.907 & 0.745 \\
StarLightCurves & \textbf{0.979} & 0.969 & 0.972 & 0.964 & 0.968 & 0.967 & 0.949 \\
Strawberry & \textbf{0.978} & 0.962 & 0.970 & 0.954 & 0.951 & 0.965 & 0.916 \\
SwedishLeaf & \textbf{0.962} & 0.941 & 0.938 & 0.914 & 0.880 & 0.923 & 0.738 \\
Symbols & 0.971 & \textbf{0.976} & 0.961 & 0.963 & 0.885 & 0.916 & 0.786 \\
SyntheticControl & \textbf{1.000} & 0.997 & 0.993 & 0.987 & \textbf{1.000} & 0.990 & 0.490 \\
ToeSegmentation1 & \textbf{0.947} & 0.917 & 0.890 & 0.939 & 0.864 & 0.930 & 0.807 \\
ToeSegmentation2 & \textbf{0.908} & 0.892 & \textbf{0.908} & 0.900 & 0.831 & 0.877 & 0.615 \\
Trace & \textbf{1.000} & \textbf{1.000} & \textbf{1.000} & 0.990 & \textbf{1.000} & \textbf{1.000} & \textbf{1.000} \\
TwoLeadECG & \textbf{0.999} & 0.986 & 0.985 & \textbf{0.999} & 0.993 & 0.976 & 0.871 \\
TwoPatterns & \textbf{1.000} & \textbf{1.000} & 0.994 & 0.999 & \textbf{1.000} & 0.999 & 0.466 \\
UMD & \textbf{1.000} & \textbf{1.000} & \textbf{1.000} & 0.993 & 0.993 & 0.986 & 0.910 \\
UWaveGestureLibraryAll & 0.878 & 0.930 & \textbf{0.956} & 0.896 & 0.903 & 0.692 & 0.475 \\
UWaveGestureLibraryX & \textbf{0.823} & 0.795 & 0.814 & 0.785 & 0.781 & 0.733 & 0.569 \\
UWaveGestureLibraryY & \textbf{0.762} & 0.719 & 0.736 & 0.710 & 0.697 & 0.641 & 0.348 \\
UWaveGestureLibraryZ & 0.769 & \textbf{0.770} & 0.749 & 0.757 & 0.721 & 0.690 & 0.655 \\
Wafer & \textbf{0.998} & \textbf{0.998} & 0.996 & 0.992 & 0.994 & 0.994 & 0.991 \\
Wine & \textbf{0.944} & 0.870 & 0.907 & 0.815 & 0.759 & 0.778 & 0.500 \\
WordSynonyms & 0.699 & 0.676 & \textbf{0.705} & 0.691 & 0.630 & 0.531 & 0.422 \\
Worms & \textbf{0.792} & 0.701 & 0.779 & 0.727 & 0.623 & 0.753 & 0.455 \\
WormsTwoClass & \textbf{0.844} & 0.805 & 0.792 & 0.792 & 0.727 & 0.753 & 0.584 \\
Yoga & 0.872 & \textbf{0.887} & 0.834 & 0.837 & 0.812 & 0.791 & 0.830 \\
\end{xltabular}

\clearpage
}

\begin{table}
\centering
\caption{Full results for time-series classification on the \textbf{UEA} archive using accuracy as metric. The reported scores for all comparison methods are taken from~\cite{ts2vec}. Note that for each method, pre-training and the downstream task are performed for each dataset individually.} \vspace{.5\baselineskip}
\label{tab:uea}
\begin{tabular}{lcccccc}
\toprule
 UEA Dataset & Ours & TS2Vec & T-Loss & TNC & TS-TCC & TST 
  \\
 \midrule
Avg. acc. & \textbf{0.738} & 0.704 & 0.658 & 0.670 & 0.668 & 0.617 \\
Avg. rank & \textbf{1.700} & 2.733 & 3.500 & 4.033 & 3.833 & 4.633 \\
\midrule
ArticularyWordRecognition & \textbf{0.990} & 0.987 & 0.943 & 0.973 & 0.953 & 0.977 \\
AtrialFibrillation & \textbf{0.267} & 0.200 & 0.133 & 0.133 & \textbf{0.267} & 0.067 \\
BasicMotions & 0.925 & 0.975 & \textbf{1.000} & 0.975 & \textbf{1.000} & 0.975 \\
CharacterTrajectories & 0.994 & \textbf{0.995} & 0.993 & 0.967 & 0.985 & 0.975 \\
Cricket & \textbf{1.000} & 0.972 & 0.972 & 0.958 & 0.917 & \textbf{1.000} \\
DuckDuckGeese & 0.480 & \textbf{0.680} & 0.650 & 0.460 & 0.380 & 0.620 \\
EigenWorms & \textbf{0.931} & 0.847 & 0.840 & 0.840 & 0.779 & 0.748 \\
Epilepsy & \textbf{0.986} & 0.964 & 0.971 & 0.957 & 0.957 & 0.949 \\
Ering & \textbf{0.919} & 0.874 & 0.133 & 0.852 & 0.904 & 0.874 \\
EthanolConcentration & \textbf{0.460} & 0.308 & 0.205 & 0.297 & 0.285 & 0.262 \\
FaceDetection & 0.541 & 0.501 & 0.513 & 0.536 & \textbf{0.544} & 0.534 \\
FingerMovements & \textbf{0.590} & 0.480 & 0.580 & 0.470 & 0.460 & 0.560 \\
HandMovementDirection & \textbf{0.432} & 0.338 & 0.351 & 0.324 & 0.243 & 0.243 \\
Handwriting & 0.428 & \textbf{0.515} & 0.451 & 0.249 & 0.498 & 0.225 \\
Heartbeat & \textbf{0.751} & 0.683 & 0.741 & 0.746 & \textbf{0.751} & 0.746 \\
InsectWingbeat & 0.449 & 0.466 & 0.156 & \textbf{0.469} & 0.264 & 0.105 \\
JapaneseVowels & 0.978 & 0.984 & \textbf{0.989} & 0.978 & 0.930 & 0.978 \\
LSST & \textbf{0.640} & 0.537 & 0.509 & 0.595 & 0.474 & 0.408 \\
Libras & \textbf{0.900} & 0.867 & 0.883 & 0.817 & 0.822 & 0.656 \\
MotorImagery & 0.520 & 0.510 & 0.580 & 0.500 & \textbf{0.610} & 0.500 \\
NATOPS & \textbf{0.972} & 0.928 & 0.917 & 0.911 & 0.822 & 0.850 \\
PEMS-SF & \textbf{0.884} & 0.682 & 0.676 & 0.699 & 0.734 & 0.740 \\
PenDigits & 0.987 & \textbf{0.989} & 0.981 & 0.979 & 0.974 & 0.560 \\
PhonemeSpectra & \textbf{0.292} & 0.233 & 0.222 & 0.207 & 0.252 & 0.085 \\
RacketSports & \textbf{0.908} & 0.855 & 0.855 & 0.776 & 0.816 & 0.809 \\
SelfRegulationSCP1 & \textbf{0.860} & 0.812 & 0.843 & 0.799 & 0.823 & 0.754 \\
SelfRegulationSCP2 & \textbf{0.600} & 0.578 & 0.539 & 0.550 & 0.533 & 0.550 \\
SpokenArabicDigits & \textbf{0.992} & 0.988 & 0.905 & 0.934 & 0.970 & 0.923 \\
StandWalkJump & \textbf{0.533} & 0.467 & 0.333 & 0.400 & 0.333 & 0.267 \\
UWaveGestureLibrary & \textbf{0.919} & 0.906 & 0.875 & 0.759 & 0.753 & 0.575 \\

\bottomrule
\end{tabular}
\end{table}

\clearpage

\begin{landscape}
\begin{table}[t]
\centering
\caption{Full results for \textbf{univariate} time-series forecasting using MSE and MAE as metrics. The reported scores for all comparison methods are taken from~\cite{ts2vec}. Note that for each method, pre-training and the downstream task are performed for each dataset individually.} \vspace{.5\baselineskip}
\label{tab:forecast_uni}
\begin{tabular}{lccccccccccccccc}
\toprule
& & \multicolumn{2}{c}{Ours} & \multicolumn{2}{c}{TS2Vec \cite{ts2vec}} & \multicolumn{2}{c}{Informer \cite{ett_informer}} & \multicolumn{2}{c}{LogTrans \cite{logtrans}} & \multicolumn{2}{c}{N-BEATS \cite{nbeats}} & \multicolumn{2}{c}{TCN \cite{tcn}} & \multicolumn{2}{c}{LSTnet \cite{lstnet}} \\
 \cmidrule(lr){3-4}  \cmidrule(lr){5-6}  \cmidrule(lr){7-8}  \cmidrule(lr){9-10}  \cmidrule(lr){11-12}  \cmidrule(lr){13-14}  \cmidrule(lr){15-16}
Dataset & $H$ & MSE & MAE  & MSE & MAE  & MSE & MAE  & MSE & MAE  & MSE & MAE  & MSE & MAE & MSE & MAE \\
\midrule
& 24 & 0.049& 0.169& \textbf{0.039} & \textbf{0.152} & 0.098& 0.247& 0.103& 0.259& 0.094& 0.238& 0.075& 0.21& 0.108& 0.284\\
& 48 & 0.079& 0.215& \textbf{0.062} & \textbf{0.191} & 0.158& 0.319& 0.167& 0.328& 0.21& 0.367& 0.227& 0.402& 0.175& 0.424\\
ETTh$_1$ & 168 & 0.136& 0.287& \textbf{0.134} & \textbf{0.282} & 0.183& 0.346& 0.207& 0.375& 0.232& 0.391& 0.316& 0.493& 0.396& 0.504\\
& 336 & 0.158& 0.315& \textbf{0.154} & \textbf{0.31} & 0.222& 0.387& 0.23& 0.398& 0.232& 0.388& 0.306& 0.495& 0.468& 0.593\\
& 720 & 0.23& 0.387& \textbf{0.163} & \textbf{0.327} & 0.269& 0.435& 0.273& 0.463& 0.322& 0.49& 0.39& 0.557& 0.659& 0.766\\
\midrule
& 24 & \textbf{0.077} & \textbf{0.21} & 0.09& 0.229& 0.093& 0.24& 0.102& 0.255& 0.198& 0.345& 0.103& 0.249& 3.554& 0.445\\
& 48 & \textbf{0.102} & \textbf{0.247} & 0.124& 0.273& 0.155& 0.314& 0.169& 0.348& 0.234& 0.386& 0.142& 0.29& 3.19& 0.474\\
ETTh$_2$ & 168 & \textbf{0.153} & \textbf{0.309} & 0.208& 0.36& 0.232& 0.389& 0.246& 0.422& 0.331& 0.453& 0.227& 0.376& 2.8& 0.595\\
& 336 & \textbf{0.183} & \textbf{0.339} & 0.213& 0.369& 0.263& 0.417& 0.267& 0.437& 0.431& 0.508& 0.296& 0.43& 2.753& 0.738\\
& 720 & \textbf{0.207} & \textbf{0.368} & 0.214& 0.374& 0.277& 0.431& 0.303& 0.493& 0.437& 0.517& 0.325& 0.463& 2.878& 1.044\\
\midrule
& 24 & 0.017& 0.098& \textbf{0.015} & \textbf{0.092} & 0.03& 0.137& 0.065& 0.202& 0.054& 0.184& 0.041& 0.157& 0.09& 0.206\\
& 48 & 0.031& 0.132& \textbf{0.027} & \textbf{0.126} & 0.069& 0.203& 0.078& 0.22& 0.19& 0.361& 0.101& 0.257& 0.179& 0.306\\
ETTm$_1$ & 96 & 0.052& 0.174& \textbf{0.044} & \textbf{0.161} & 0.194& 0.372& 0.199& 0.386& 0.183& 0.353& 0.142& 0.311& 0.272& 0.399\\
& 288 & 0.109& 0.257& \textbf{0.103} & \textbf{0.246} & 0.401& 0.554& 0.411& 0.572& 0.186& 0.362& 0.318& 0.472& 0.462& 0.558\\
& 672 & 0.162& 0.316& \textbf{0.156} & \textbf{0.307} & 0.512& 0.644& 0.598& 0.702& 0.197& 0.368& 0.397& 0.547& 0.639& 0.697\\
\midrule
& 24 & 0.272& 0.386& 0.26& 0.288& \textbf{0.251} & \textbf{0.275} & 0.528& 0.447& 0.427& 0.33& 0.263& 0.279& 0.281& 0.287\\
& 48 & \textbf{0.303} & 0.402& 0.319& \textbf{0.324} & 0.346& 0.339& 0.409& 0.414& 0.551& 0.392& 0.373& 0.344& 0.381& 0.366\\
Electricity & 168 & \textbf{0.337} & 0.431& 0.427& \textbf{0.394} & 0.544& 0.424& 0.959& 0.612& 0.893& 0.538& 0.609& 0.462& 0.599& 0.5\\
& 336 & \textbf{0.36} & \textbf{0.449} & 0.565& 0.474& 0.713& 0.512& 1.079& 0.639& 1.035& 0.669& 0.855& 0.606& 0.823& 0.624\\
& 720 & \textbf{0.359} & \textbf{0.453} & 0.861& 0.643& 1.182& 0.806& 1.001& 0.714& 1.548& 0.881& 1.263& 0.858& 1.278& 0.906\\
\midrule
Avg. & & \textbf{0.169} & 0.297& 0.209& \textbf{0.296} & 0.31& 0.39& 0.37& 0.434& 0.399& 0.426& 0.338& 0.413& 1.099& 0.536 \\
\bottomrule
\end{tabular}
\end{table}
\end{landscape}

\clearpage

\begin{landscape}
\begin{table}[t]
\centering
\caption{Full results for \textbf{multivariate} time-series forecasting using MSE and MAE as metrics. The reported scores for all comparison methods are taken from~\cite{ts2vec}. Note that for each method, pre-training and the downstream task are performed for each dataset individually.} \vspace{.5\baselineskip}
\label{tab:forecast_multi}
\begin{tabular}{lccccccccccccc}
\toprule
& & \multicolumn{2}{c}{Ours} & \multicolumn{2}{c}{TS2Vec \cite{ts2vec}} & \multicolumn{2}{c}{Informer \cite{ett_informer}} & \multicolumn{2}{c}{TCN \cite{tcn}} & \multicolumn{2}{c}{LogTrans \cite{logtrans}} & \multicolumn{2}{c}{LSTnet \cite{lstnet}} \\
\cmidrule(lr){3-4}  \cmidrule(lr){5-6}  \cmidrule(lr){7-8}  \cmidrule(lr){9-10}  \cmidrule(lr){11-12}  \cmidrule(lr){13-14}
Dataset & $H$ & MSE & MAE  & MSE & MAE  & MSE & MAE  & MSE & MAE  & MSE & MAE  & MSE & MAE \\
\midrule
& 24 & \textbf{0.504} & \textbf{0.518} & 0.599& 0.534& 0.577& 0.549& 0.767& 0.612& 0.686& 0.604& 1.293& 0.901\\
& 48 & \textbf{0.547} & \textbf{0.548} & 0.629& 0.555& 0.685& 0.625& 0.713& 0.617& 0.766& 0.757& 1.456& 0.96\\
ETTh$_1$ & 168 & 0.689& 0.636& 0.755& 0.636& 0.931& 0.752& 0.995& 0.738& 1.002& 0.846& 1.997& 1.214\\
& 336 & \textbf{0.78} & \textbf{0.687} & 0.907& 0.717& 1.128& 0.873& 1.175& 0.8& 1.362& 0.952& 2.655& 1.369\\
& 720 & \textbf{0.813} & \textbf{0.69} & 1.048& 0.79& 1.215& 0.896& 1.453& 1.311& 1.397& 1.291& 2.143& 1.38\\
\midrule
& 24 & \textbf{0.293} & \textbf{0.409} & 0.398& 0.461& 0.72& 0.665& 1.365& 0.888& 0.828& 0.75& 2.742& 1.457\\
& 48 & \textbf{0.412} & \textbf{0.493} & 0.58& 0.573& 1.457& 1.001& 1.395& 0.96& 1.806& 1.034& 3.567& 1.687\\
ETTh$_2$ & 168 & \textbf{0.86} & \textbf{0.737} & 1.901& 1.065& 3.489& 1.515& 3.166& 1.407& 4.07& 1.681& 3.242& 2.513\\
& 336 & \textbf{1.003} & \textbf{0.805} & 2.304& 1.215& 2.723& 1.34& 3.256& 1.481& 3.875& 1.763& 2.544& 2.591\\
& 720 & \textbf{1.013} & \textbf{0.803} & 2.65& 1.373& 3.467& 1.473& 3.69& 1.588& 3.913& 1.552& 4.625& 3.709\\
\midrule
& 24 & 0.342& 0.41& 0.443& 0.436& \textbf{0.323} & \textbf{0.369} & 0.324& 0.374& 0.419& 0.412& 1.968& 1.17\\
& 48 & \textbf{0.438} & 0.48& 0.582& 0.515& 0.494& 0.503& 0.477& \textbf{0.45} & 0.507& 0.583& 1.999& 1.215\\
ETTm$_1$ & 96 & \textbf{0.482} & \textbf{0.511} & 0.622& 0.549& 0.678& 0.614& 0.636& 0.602& 0.768& 0.792& 2.762& 1.542\\
& 288 & \textbf{0.584} & \textbf{0.576} & 0.709& 0.609& 1.056& 0.786& 1.27& 1.351& 1.462& 1.32& 1.257& 2.076\\
& 672 & \textbf{0.685} & \textbf{0.634} & 0.786& 0.655& 1.192& 0.926& 1.381& 1.467& 1.669& 1.461& 1.917& 2.941\\
\midrule
& 24 & \textbf{0.274} & 0.375& 0.287& \textbf{0.374} & 0.312& 0.387& 0.305& 0.384& 0.297& \textbf{0.374} & 0.356& 0.419\\
& 48 & \textbf{0.288} & \textbf{0.384} & 0.307& 0.388& 0.392& 0.431& 0.317& 0.392& 0.316& 0.389& 0.429& 0.456\\
Electricity & 168 & \textbf{0.301} & \textbf{0.393} & 0.332& 0.407& 0.515& 0.509& 0.358& 0.423& 0.426& 0.466& 0.372& 0.425\\
& 336 & \textbf{0.305} & \textbf{0.398} & 0.349& 0.42& 0.759& 0.625& 0.349& 0.416& 0.365& 0.417& 0.352& 0.409\\
& 720 & \textbf{0.317} & 0.409& 0.375& 0.438& 0.969& 0.788& 0.447& 0.486& 0.344& \textbf{0.403} & 0.38& 0.443\\
\midrule
Avg. & & \textbf{0.546} & \textbf{0.545} & 0.828& 0.635& 1.154& 0.781& 1.192& 0.837& 1.314& 0.892& 1.903& 1.444\\
\bottomrule
\end{tabular}
\end{table}
\end{landscape}

\end{document}